\documentclass[pdflatex,sn-nature]{sn-jnl}

\usepackage{graphicx}%
\usepackage{multirow}%
\usepackage{amsmath,amssymb,amsfonts}%
\usepackage{amsthm}%
\usepackage{mathrsfs}%
\usepackage[title]{appendix}%
\usepackage{xcolor}%
\usepackage{textcomp}%
\usepackage{manyfoot}%
\usepackage{booktabs}%
\usepackage{algorithm}%
\usepackage{algorithmicx}%
\usepackage{algpseudocode}%
\usepackage{listings}%

\theoremstyle{thmstyleone}%
\theoremstyle{thmstyletwo}%

\theoremstyle{thmstylethree}%

\begin{document}

\title[Neural emulation of gravity-driven geohazard runout]{Neural emulation of gravity-driven geohazard runout}


\author*[1,2]{\fnm{Lorenzo} \sur{Nava}}\email{ln413@cam.ac.uk}

\author[1,3]{\fnm{Ye} \sur{Chen}}\email{tj\_chenye@tongji.edu.cn}

\author[1,2,4]{\fnm{Maximillian} \sur{Van Wyk de Vries}}\email{msv27@cam.ac.uk}

\affil*[1]{\orgdiv{Department of Earth Sciences}, \orgname{University of Cambridge}, \orgaddress{\street{Downing St}, \city{Cambridge}, \postcode{CB2 3EQ}, \country{United Kingdom}}}

\affil[2]{\orgdiv{Department of Geography}, \orgname{University of Cambridge}, \orgaddress{\street{Downing PI}, \city{Cambridge}, \postcode{CB2 1DB}, \country{United Kingdom}}}

\affil[3]{\orgdiv{Department of Geotechnical Engineering}, \orgname{Tongji University}, \orgaddress{\street{Siping Rd}, \city{Shanghai}, \postcode{200092}, \country{China}}}

\affil[4]{\orgdiv{Scott Polar Research Institute}, \orgname{University of Cambridge}, \orgaddress{\street{Lensfield Rd}, \city{Cambridge}, \postcode{CB2 1ER}, \country{United Kingdom}}}


\abstract{Predicting geohazard runout is critical for protecting lives, infrastructure and ecosystems.
Rapid mass flows, including landslides and avalanches, cause several thousand deaths across a wide range of environments, often travelling many kilometres from their source. The wide range of source conditions and material properties governing these flows makes their runout difficult to anticipate, particularly for downstream communities that may be suddenly exposed to severe impacts.
Accurately predicting runout at scale requires models that are both physically realistic and computationally efficient, yet existing approaches face a fundamental speed–realism trade-off. 
Here we train a machine learning model to predict geohazard runout across representative real-world terrains. 
The model predicts both flow extent and deposit thickness with high accuracy and $10^{2}$--$10^{4}\times$ faster computation than numerical solvers. It is trained over  $10^{5}$ numerical simulations across  $10^{4}$ real-world digital elevation model chips and reproduces key physical behaviours, including avulsion and deposition patterns, while generalizing across different flow types, sizes and landscapes.
Our results demonstrate that neural emulation enables rapid, spatially resolved runout prediction across diverse real-world terrains, opening new opportunities for disaster risk reduction and impact-based forecasting. These results highlight neural emulation as a promising pathway for extending physically realistic geohazard modelling to spatial and temporal scales relevant for large-scale early-warning systems.}




\maketitle

\section*{Main}\label{sec1}

Rapid geophysical flows such as landslides, avalanches, and volcanic flows are among the most destructive natural hazards on Earth \citep{Froude2018, Lacroix2020, Eckert2024, de2015landslides}. Climate change is increasing the frequency and intensity of extreme precipitation, snowmelt, and storm events, modifying the likelihood of many rapid mass movements \citep{stanley2024landslide}. At the same time, accelerating urbanization and infrastructure expansion are increasing exposure to these hazards \citep{Ozturk2022climate}.
These flows can travel many kilometres from their initiation zone, threatening lives, infrastructure, and ecosystems \citep{Shugar2021}. Predicting how far they will flow, their runout, is central to hazard assessment, land-use planning, and impact-based forecasting. Runout defines the true impact zone and determines the magnitude of damage once failure occurs. Recent analyses suggest that landslides alone affect more than 17\% of the world’s land area, with roughly 8\% of the global population living within potential runout zones \citep{jia2021global}. Yet anticipating the mobility and spatial footprint of these flows remains challenging, particularly when forecasts are needed fast enough to support operational decisions \citep{yanites2025cascading}. 

In forecasting and hazard assessment systems, this computational delay is critical: emergency agencies need runout estimates quickly enough to guide evacuations and prioritise response, not hours after a triggering event \citep{world2012guidelines}. The need for speed becomes even more acute at regional scales, where thousands of potential initiation zones must be screened rapidly before or during intense rainfall or following an earthquake. These time constraints also restrict our ability to generate probabilistic forecasts from large ensembles, which is an essential component of robust, uncertainty-aware decision-making during hazardous events \citep{krzysztofowicz2001case, canli2018probabilistic}.

Existing runout models fall into two broad classes, each with inherent trade-offs. Empirical and analytical approaches use geometric correlations to estimate runout length or area rapidly, but they neglect the physics of flow motion and cannot reproduce key behaviours such as avulsion or deposition \citep{mcdougall20172014}. Physics-based numerical models (e.g., VolcFlow \citep{kelfoun2005numerical}, and \textit{r.avaflow} \citep{mergili2017r}) resolve depth, velocity, and impact pressure, offering physically realistic insights into flow dynamics. However, they are computationally intensive, even requiring hours for a single simulation, and depend on poorly constrained inputs such as rheology and initial conditions \citep{mcdougall20172014}. As a result, their use is largely restricted to retrospective, site-specific studies. Statistical emulators exist but are generally limited to basin-scale applications \citep{zhao2021emulator}. At regional or global scales, this speed–realism trade-off prevents large ensemble analyses and limits our ability to propagate uncertainty \citep{kabir2018neural} and provide probabilistic predictions, critical requirements for operational risk assessment and forecasting systems \citep{farmer2017uncertainty}.

\begin{figure}[!t]
\centering
\label{fig:scheme}
\includegraphics[width=\textwidth]{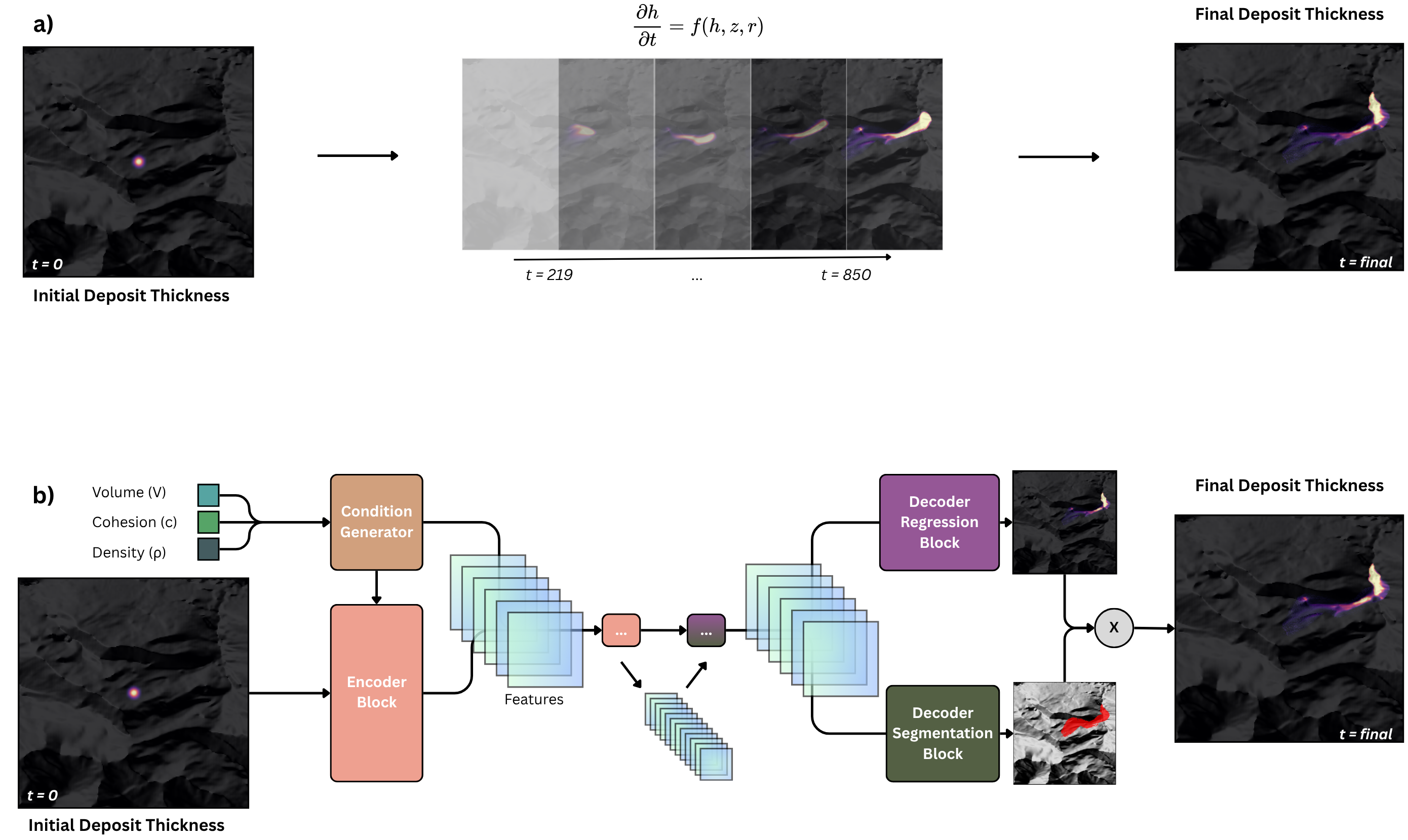}
\caption{Comparison between the structure of a physics-based granular flow model (a) and our geophysical flow neural network emulator (b).}
\end{figure}

Here we present a physics-based data-driven framework that bridges the divide between empirical speed and physical realism in geohazard runout prediction. We develop a neural emulator trained on tens of thousands of physics-based simulations to reproduce flow runout across diverse real-world terrains and material properties. The model predicts flow extent and deposit thickness with high accuracy while operating up to four orders of magnitude faster than conventional solvers. Our results demonstrate that spatially constrained, gravity-driven mass flows can be emulated universally across varied landscapes and rheologies. This work lays the groundwork for real-time, physically grounded hazard forecasting and provides a foundation for future multi-hazard extensions.

\section*{Combining neural networks with physical simulations}

We model the behaviour of rapid, gravity-driven mass flows, such as landslides, and avalanches, using a neural network trained on physics-based numerical simulations.   
Each simulation initializes a Gaussian-shaped pile of material on a 30 m Copernicus Global Digital Elevation Model (COP30) tile and evolves it downslope under gravity using a depth-averaged \citep{kelfoun2005numerical}, frictional rheology \citep{Bartelt_Salm_Gruber_1999}.  
The reference numerical model solves the depth-averaged mass- and momentum-conservation equations with a turbulent basal friction law (Voellmy model) and variable material properties (more details in Section \ref{sec:numerical}). Bulk density~(\(\rho\)) ranges from 917 to 2650~\(\mathrm{kg\,m^{-3}}\), cohesion~(\(c\)) from 5 to 50~\(\mathrm{kPa}\), and initial volume~(\(V\)) from \(10^{4}\) to \(10^{7}\)~\(\mathrm{m^{3}}\), spanning the transition from ice-rich to rock-dominated materials and from small to large-scale events.

Each simulation produces two target maps: the final runout footprint and the spatial distribution of deposit thickness.  
We treat the numerical solver as the physical reference and evaluate whether a neural network can reproduce these outputs directly from real topography and bulk flow parameters, generalizing across diverse terrains worldwide.
 
Formally, the task is to approximate the mapping
\[
\mathcal{F} : \{\mathbf{X}_{\mathrm{topo}},\, V,\, \rho,\, c\}
\;\longrightarrow\;
\{\hat{M},\, \hat{h}(x,y)\},
\]
where \(\mathbf{X}_{\mathrm{topo}} \in \mathbb{R}^{H \times W \times C}\) denotes a stack of \(C\) topographic raster channels derived from the digital elevation model, \(V\) is the source volume, \(\rho\) the bulk density, and \(c\) the cohesion. The outputs \(\hat{M}\) and \(\hat{h}(x,y)\) denote the predicted runout footprint and final deposit thickness field, respectively.

The independent test set comprises 9{,}279 numerical simulations generated on unseen DEM tiles randomly sampled from several major mountain ranges worldwide, with parameter combinations spanning the full range of volumes, densities, and cohesions used in training.  
Our goal is to evaluate both predictive fidelity, how closely the emulator reproduces numerical solutions, and computational efficiency, demonstrating that comparable accuracy can be achieved at orders-of-magnitude lower cost.

\subsection*{Proof of concept on synthetic data}

\begin{figure}[!t]
\label{fig:compa1}
\centering
\includegraphics[width=\textwidth]{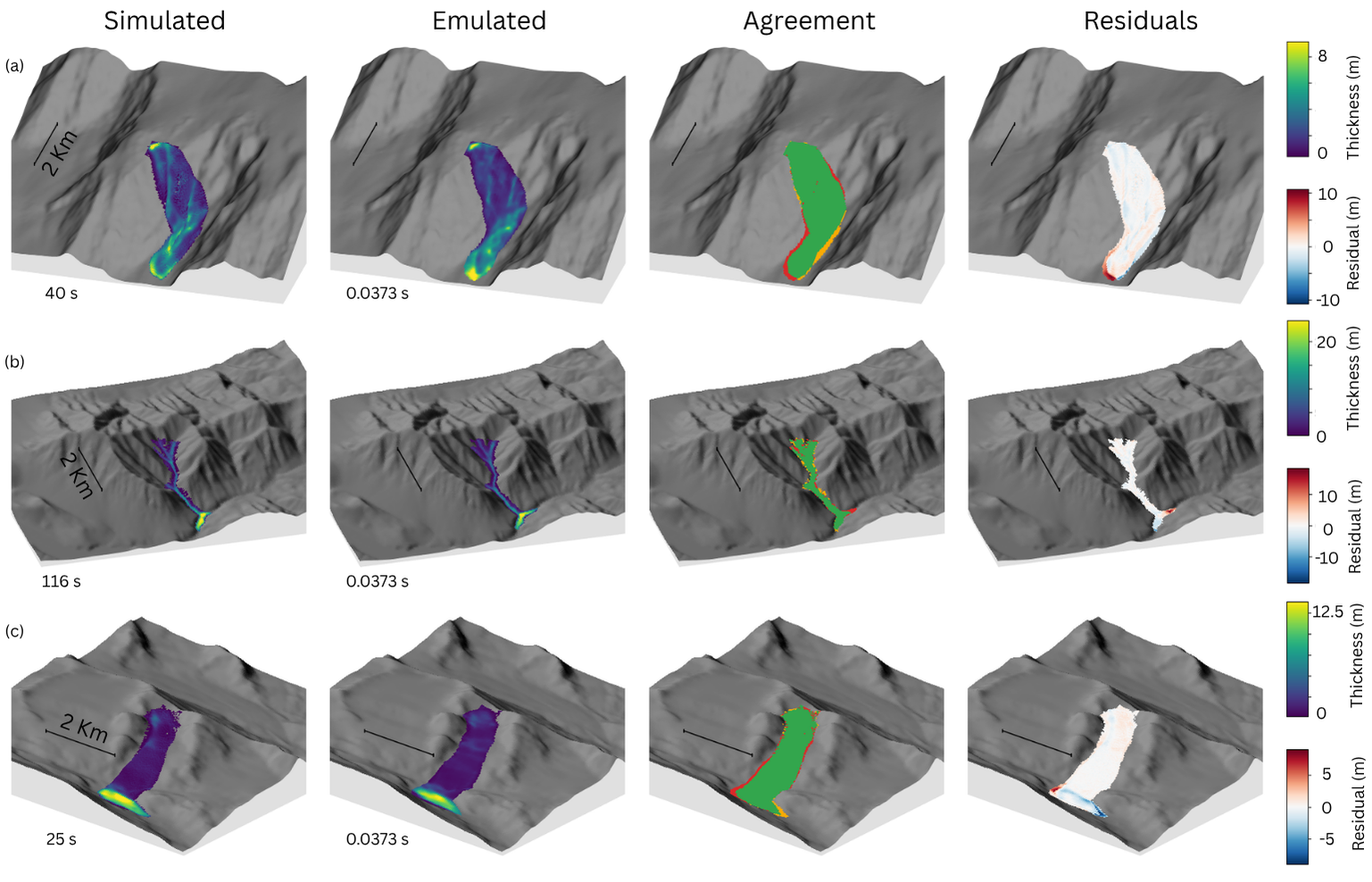}
\caption{%
Comparison between numerical simulations and neural emulator predictions for three representative test cases. 
Each row corresponds to a different flow scenario with varying topography and material properties. 
Columns show (from left to right): simulated deposit thickness, emulated deposit thickness, binary agreement between simulated and predicted runout footprints, and residuals in deposit thickness. 
The emulator reproduces the spatial extent and thickness patterns of the numerical solutions with high fidelity, capturing key flow features such as channelized propagation and lateral spreading while reducing computation time from tens of seconds to $\sim$0.04~s per case. 
(a) $V = 9.7\times10^{6}$~m$^{3}$, $c = 21.3$~kPa, $\rho = 1477$~kg~m$^{-3}$; 
(b) $V = 5.7\times10^{6}$~m$^{3}$, $c = 19.7$~kPa, $\rho = 2092$~kg~m$^{-3}$; 
(c) $V = 9.6\times10^{6}$~m$^{3}$, $c = 6.2$~kPa, $\rho = 1391$~kg~m$^{-3}$.%
}
\end{figure}

\begin{figure}[!t]
\label{fig:fig3}
\centering
\includegraphics[width=\textwidth]{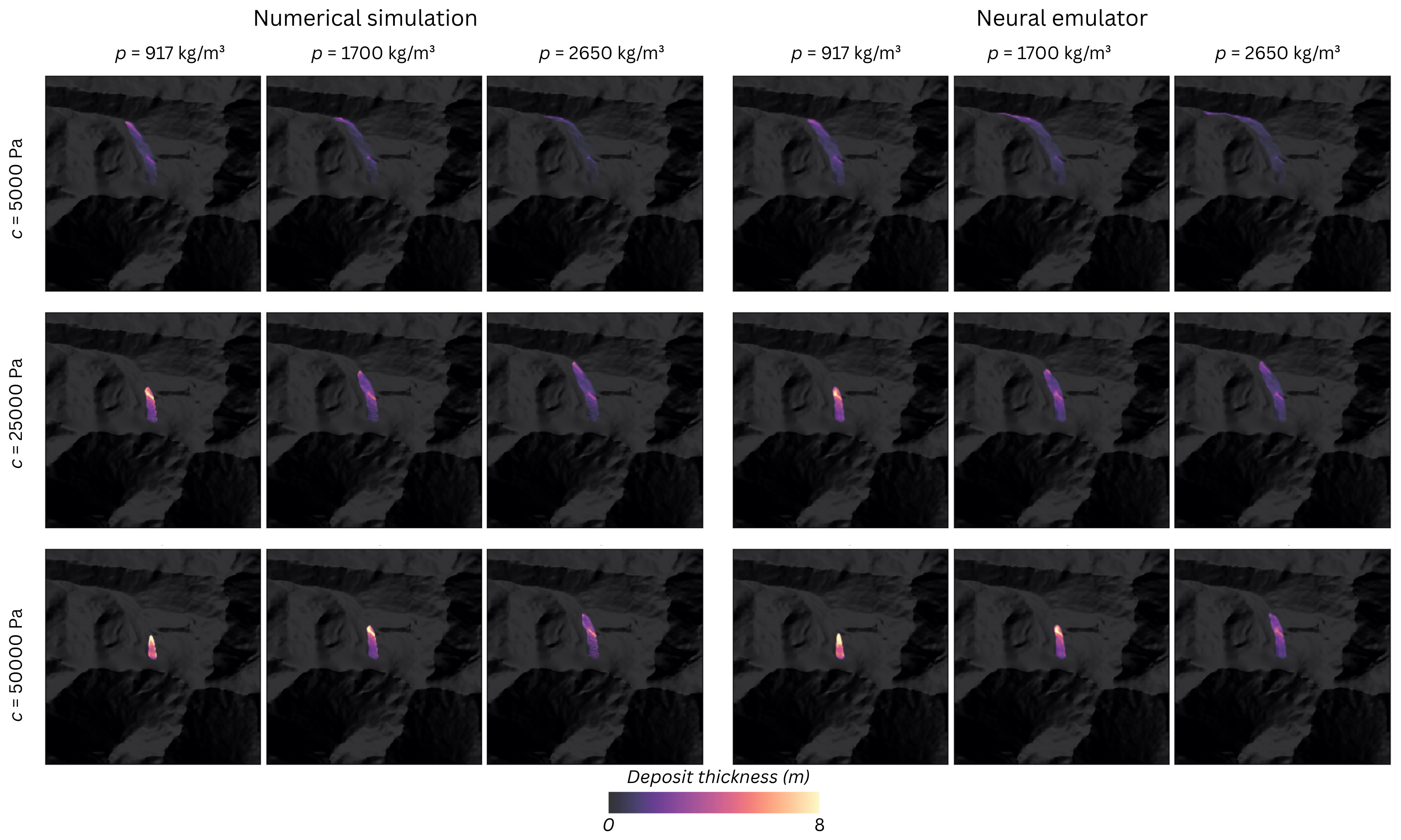}
\caption{%
Comparison between numerical simulations (left panels) and neural network predictions (right panels) for systematic variations in material properties on a fixed topography. 
Rows correspond to increasing cohesion ($c = 5$, $25$, and $50$~kPa), and columns to increasing bulk density ($\rho = 917$, $1700$, and $2650$~kg~m$^{-3}$). 
Both models reproduce the expected decrease in mobility with increasing cohesion and decreasing density, with consistent spatial patterns of deposition. 
Minor overprediction occurs for the highest densities, where nonlinear rheological effects are strongest, but overall the emulator captures the coupled influence of $\rho$ and $c$ on flow extent and deposit thickness with high fidelity. %
}

\end{figure}

The model reproduces numerical runout simulations with high accuracy across the independent test set composed of unseen DEMs. Predicted runout footprints achieve a mean intersection-over-union (IoU) of \textbf{0.84} and an F\(_1\) score of \textbf{0.91}, while deposit-thickness fields yield a mean root-mean-square error (RMSE) of \textbf{1.6 m} within inundated areas. Predicted maximum runout distances closely match numerical solutions, with a mean of 26.05 ± 25.04 pixels compared to 25.96 ± 24.90 pixels for the reference simulations. The mean absolute error in maximum runout distance is 1.3 pixels ($\approx 40 m$ at 30 m resolution).
Errors remain low across the sampled parameter space, increasing moderately with flow volume and more strongly with bulk density.  
The latter trend reflects reduced accuracy for the most nonlinear rheologies, where FiLM conditioning captures but underestimates density-driven mobility changes (Fig. \ref{fig:fig3}).  
Despite these variations, the emulator preserves key physical behaviours, including channelized propagation, lateral avulsion, and spatially coherent deposition patterns across diverse terrains (Fig. \ref{fig:compa1}).  
Overall, the emulator provides physically consistent and quantitatively accurate representations of gravity-driven, spatially constrained flows across the full range of volumes, densities, and cohesions represented in the training data.  

In addition to accuracy, the neural emulator delivers substantial computational gains.  A single forward pass for a $256 \times 256$ tile requires only 0.04 s on a single GPU, compared with 10–200~s for the equivalent numerical simulation on a single CPU, both executed on the same desktop machine.  
This yields a speedup of about \(10^{3}\)–\(10^{4}\). Batch inference scales linearly, allowing ensembles of over 1{,}000 parameter realizations to be evaluated in \(\sim 90\,\text{s}\) on a single GPU. This efficiency enables probabilistic forecasting and large-scale uncertainty analyses that are impractical with conventional numerical solvers.

\subsection*{Probabilistic forecasting and uncertainty propagation}

The large computational gains achieved by the emulator enable full uncertainty propagation within seconds for each scenario. This is crucial for operational hazard assessment over large areas, where initial conditions, such as source volume, density, or cohesion, are often poorly constrained but strongly influence runout outcomes.  
In practical forecasting and risk assessment, these parameters can be treated probabilistically, conditioned either on susceptibility estimates or on real-time ground motion observations that indicate an impending failure.

We illustrate the emulator's capability for probabilistic prediction using three representative case studies: 
(a) the Zymoetz River landslide (Canada), a rock avalanche transitioning into a debris flow \citep{mcdougall2006zymoetz},
(b) a snow avalanche in the Swiss Alps \citep{hafner2021mapping}, and 
(c) the Maoxian rock avalanche (China) \citep{fan2017failure, intrieri2018maoxian}.
These examples span distinct process types, from ice-dominated to fully granular and debris-rich flows, and thus provide a strong test of model generality.  

\begin{figure}[!t]
\centering
\label{fig:fig4}
\includegraphics[width=\textwidth]{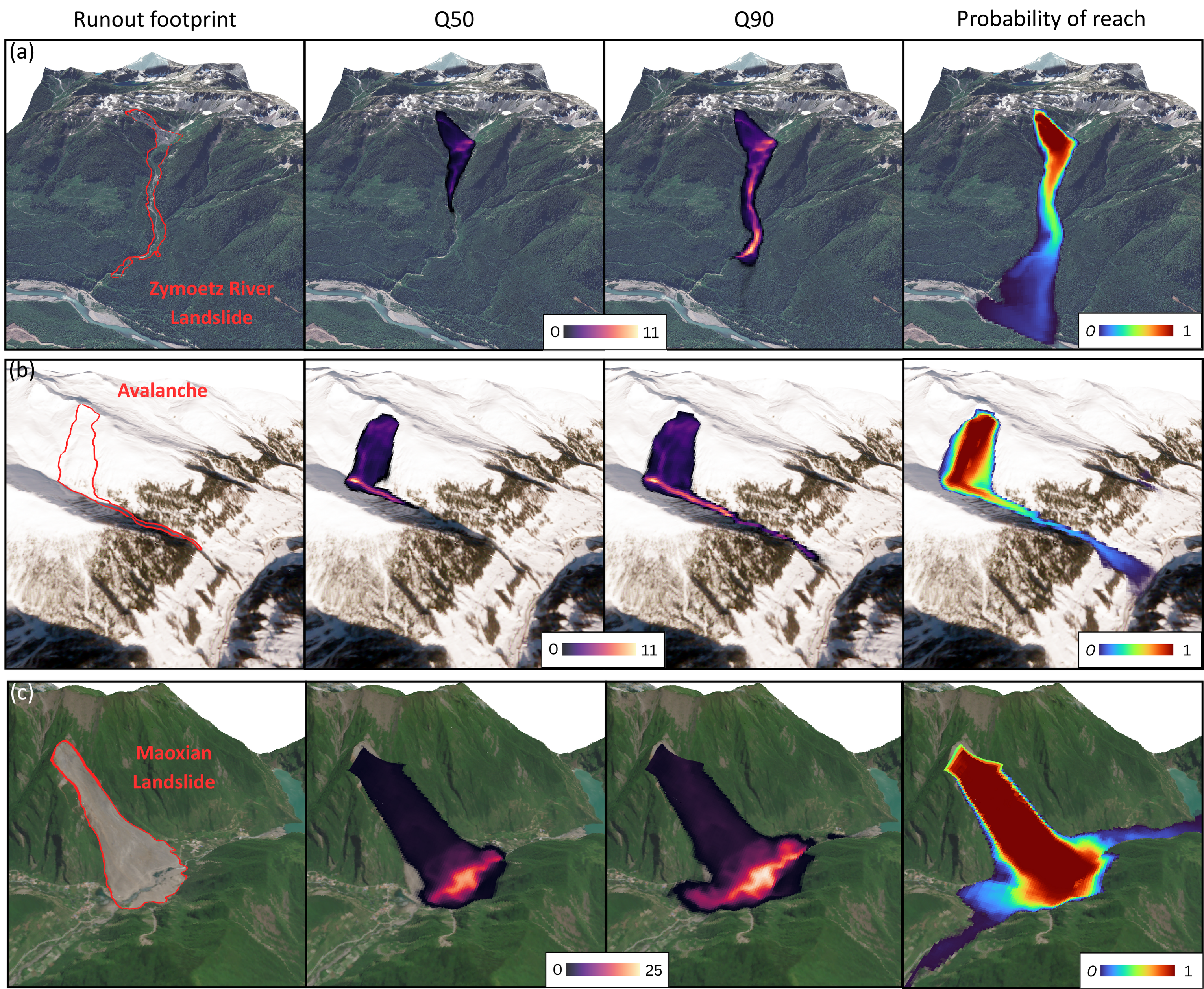}
\caption{%
Probabilistic ensembles for three distinct types of rapid mass movements, showing pixelwise probability of reach and the 50th and 90th percentile deposit-thickness quantiles derived from 1{,}024 emulator runs. 
Each ensemble explores uncertainty across key flow parameters---volume, bulk density, and cohesion---within physically plausible ranges. 
(a) \textbf{Zymoetz River landslide, Canada:} rock avalanche transitioning into debris flow, with $V = 8\times10^{5}\text{--}3\times10^{6}\,\mathrm{m^{3}}$, $\rho = 1600\text{--}2200\,\mathrm{kg\,m^{-3}}$, and $c = 5\text{--}50\,\mathrm{kPa}$. 
(b) \textbf{Swiss Alps avalanche:} snow/ice avalanche with $\rho = 917\text{--}1100\,\mathrm{kg\,m^{-3}}$, $c = 5\text{--}15\,\mathrm{kPa}$, and $V = 8\times10^{5}\text{--}3\times10^{6}\,\mathrm{m^{3}}$. 
(c) \textbf{Maoxian landslide, China:} rock avalanche with $V = 5\times10^{6}\text{--}1\times10^{7}\,\mathrm{m^{3}}$, $\rho = 1600\text{--}2400\,\mathrm{kg\,m^{-3}}$, and $c = 5\text{--}50\,\mathrm{kPa}$. 
All ensembles were computed in $90\text{--}110$~s on a single GPU. 
The maps illustrate the emulator's ability to generate spatially resolved probabilistic forecasts across diverse flow types and rheological conditions. Source: Google Satellite 2025, Copernicus Sentinel-2. %
}

\end{figure}

For each site, we define a broad parameter space encompassing realistic mobility ranges and centred on the estimated source volume. 
Input parameters were sampled using a Sobol low-discrepancy sequence to ensure uniform coverage, although any probability distribution can be specified where prior information is available.  
A total of 1{,}024 emulator runs were completed in \textbf{90~s} of wall-clock time on a single GPU.  
The resulting ensembles yield pixelwise probabilistic maps of impact, including the probability of reach and the 50th and 90th percentile deposit-thickness quantiles ($q_{50}$, $q_{90}$) (Fig.~\ref{fig:fig4}).  

Such probabilistic forecasts can be generated in near real time for multiple potential source zones, enabling large-scale, ensemble-based hazard assessments.

\section*{Implications for risk assessment and impact-based early warning}

Our results demonstrate that an appropriately designed neural network can emulate gravity-driven mass flows with accuracy comparable to numerical solvers, yet at four orders of magnitude higher computational speed. 
By reproducing runout behaviour directly from real-world topography and bulk material parameters, the emulator enables hazard assessment at spatial and temporal scales previously unattainable. 
With computation times reduced from tens of seconds to milliseconds, ensemble modelling and uncertainty propagation become operationally feasible. This allows probabilistic forecasts for entire regions in near real time. This shift has the potential to transform runout modelling from site-specific analyses into a scalable forecasting framework, providing a practical tool for both impact-based early warning and disaster risk reduction.  

The current framework generalizes across a global range of topographies, volumes, densities, and cohesions. It does not yet capture transient dynamics or processes such as entrainment, erosion, or evolving boundary conditions, and its outputs are limited to the final deposit thickness rather than the full spatiotemporal evolution of motion. 
Moreover, training at 30~m resolution on $256\times256$~pixel DEM tiles restricts direct application to higher-resolution data or events with runouts exceeding about 4~km. 
Future developments could leverage neural operator architectures to overcome these limitations, enabling scale-independent emulation and time-resolved predictions of flow evolution.  Beyond landslides, the same principle could extend to other computationally intensive Earth-surface processes governed by depth-averaged dynamics, including volcanic plumes, glacier surges, and pyroclastic density currents. 
Coupled with near-real-time monitoring and satellite data, such emulators could underpin global, probabilistic, impact-based early warning systems. They could transform static susceptibility maps into dynamic forecasts following earthquakes or rainfall events, update hazard estimates as typhoons approach, and provide seasonal probabilistic assessments of avalanche activity, all within seconds and at continental scales.

\section{Methods}
\subsection{Global synthetic dataset of geohazard runouts}

We constructed a globally distributed dataset of synthetic geohazard flows runouts by combining digital elevation models (DEMs) with randomized material source. DEMs were sampled from the Copernicus Global 30~m DEM (COP30), which provides seamless global coverage. We extracted $256 \times 256$ pixel tiles (7.68~$\times$~7.68~km) from montane regions worldwide and retained only those with mean slope $> 5^{\circ}$ in the central $11 \times 11$ pixels. To suppress striping artifacts, DEMs were smoothed with a Gaussian filter ($\sigma = 1$).  

Sources were initialized as Gaussian-shaped piles centred on each DEM tile. Gaussian-shaped piles are commonly used in runout benchmarking and provide a smooth, well-posed initial thickness distribution. Kernel size was adjusted adaptively to achieve a target mean initial thickness of 25~m (limits 7-21 pixels), covering initial source diameters from ~200 to 600 m, and piles were scaled to assigned volumes. Parameter combinations for volume, bulk density, and cohesion were generated using Latin Hypercube Sampling to ensure uniform coverage of the multidimensional parameter space. Volumes were drawn log-uniformly between $10^{4}$ and $10^{7}$~m$^{3}$. Bulk density ($\rho$) was sampled between 917 and 2650~kg~m$^{-3}$, spanning ice-rich to quartz-rich materials. Cohesion ($c$) was sampled between 5 and 50~kPa, covering weak soils to more lithified debris.

Simulations were run until maximum velocity fell below 0.5~m~s$^{-1}$. From each run we extracted the final deposit thickness field $h(x)$ at 30~m resolution and a binary footprint mask defined by $h > 0.05$~m. Runs with negligible mobility ($< 15$ displaced pixels) were discarded to exclude trivial collapses producing negligible runout. From approximately 120,000 initial simulations (ten per DEM tile), filtering yielded about 90,000 usable cases across over 12,000 unique terrain chips. The dataset was split into training (80\%), validation (10\%), and test (10\%) subsets, with no overlap in DEM tiles between splits and proportional distribution of rheologies and volumes.

\subsection{Numerical simulation}
\label{sec:numerical}
The numerical simulations were produced by using a depth-averaged flow model primarily based on \citep{kelfoun2005numerical}. The original model is designed as a bulk material sliding over a thin basal layer. This type of model has been widely applied to simulate flow-like hazards such as rock avalanches and debris flows. In this study, the model solves the following conservation equations of mass and momentum in a coordinate system referenced to the input DEM,

\begin{equation}
    \frac{\partial h}{\partial t}
    + \frac{\partial (hu)}{\partial x}
    + \frac{\partial (hv)}{\partial y} = 0
    \label{eq:mass_conservation}
\end{equation}

\begin{equation}
    \frac{\partial (hu)}{\partial t}
    + \frac{\partial (hu^2)}{\partial x}
    + \frac{\partial (huv)}{\partial y}
    = gh \sin \alpha_x
    - \frac{1}{2} \frac{\partial}{\partial x} \left( g h^2 \cos \alpha \right)
    + F_x
    \label{eq: momentum_conservation_x}
\end{equation}

\begin{equation}
    \frac{\partial (hv)}{\partial t}
    + \frac{\partial (hvu)}{\partial x}
    + \frac{\partial (hv^2)}{\partial y}
    = gh \sin \alpha_y
    - \frac{1}{2} \frac{\partial}{\partial x} \left( g h^2 \cos \alpha \right)
    + F_y
    \label{eq: momentum_conservation_y}
\end{equation}

\noindent where \textit{h} is the flow thickness, $\textbf{u} = (u,v)$ is the flow velocity, $\alpha$ is the ground slope. \textit{F} denotes the momentum source term related to friction, which is

\begin{equation}
    F_x = 
    - h \tau_{zx} / \rho
    \label{eq: friction_term}
\end{equation}

\noindent where $\rho$ is the material density. To enhance computational efficiency for a large number of simulations, we simplify the original model to retain only the most important physical concepts controlling the flow, gravity, the pressure gradient from the free surface, and basal friction, which appear as source terms in the momentum conservation.

The frictional behaviour in the simulations is represented by the Voellmy fluid model \citep{Bartelt_Salm_Gruber_1999}, a widely adopted approach developed for describing avalanche motion. This model characterise basal friction through two components: a Columb-like friction term and a Chezy-like velocity-dependent turbulent resistance term, which can be expressed as 

\begin{equation}
    \tau_{zx} = \mu \sigma_z + \frac{\rho g}{\xi} \|\mathbf{u}\|^2
    \label{eq: voellmy}
\end{equation}

\noindent where $\sigma_z$ is the overburden pressure, $\mu$ and $\xi$ are two parameters related to material properties and basal roughness. In our simulations, we used a simplified parameterisation, where the turbulent friction coefficient was fixed at $\xi$ = 0.02~m/s$^2$ and the Columb friction coefficient at $\mu$ = 0.

The flow rheology is represented by a Bingham-type formulation, in which the yield stress is controlled by a single rheological parameter, namely the cohesion \textit{c}. In the depth-averaged momentum balance, this results a resistive acceleration form as

\begin{equation}
    \mathbf{a}_{\mathrm{yield}}
    =
    - \frac{c}{\rho h}
    \frac{\mathbf{u}}{\lVert \mathbf{u} \rVert}
\label{eq: bingham}
\end{equation}

\noindent This simplified parameterisation enables the generation of a wide range of flow mobilities while keeping the number of rheological parameters to a minimum.

The numerical implementation of the model follows a staggered finite volume scheme. The conserved quantities, including mass and momentum, are updated at cell centres, while fluxes are computed across the cell interfaces. An upwind scheme is used to determine the interface fluxes depending on the sign of the velocity. These implementations improve the numerical stability and the simulation fidelity. Additional parameters are introduced for further maintaining the numerical stability. The minimum flow thickness threshold is $5 \times 10^{-4}$~m. For temporal integration, the solver starts with an initial time step of 0.25~s, which is adaptively adjusted according to the CFL condition, with a minimum allowable time step of 0.05~s.

\subsection{Neural architecture}

We adopt a U\mbox{-}Net--style encoder–decoder \citep{Ronneberger2015} with residual blocks, attention gates \citep{Oktay2018}, and feature-wise linear modulation (FiLM) \citep{perez2018film} to condition predictions on global flow parameters. The network takes as input a $C_{\mathrm{in}}$--channel raster stack ($C_{\mathrm{in}}{=}8$ in our experiments) and a low-dimensional parameter vector $\mathbf{p}\in\mathbb{R}^{d}$ ($d{=}3$; i.e., volume, density, cohesion). It outputs a runout mask (logits) and a non\mbox{-}negative deposit thickness field.  

The encoder comprises four stages, each downsampling the input by a factor of two via max pooling. Each stage uses a residual convolutional block with two $3{\times}3$ convolutions, Group Normalization with eight groups, ReLU activations, and dropout ($p{=}0.2$), together with a $1{\times}1$ projection on the skip path when channel dimensions change. GroupNorm was preferred to BatchNorm because stable normalization was essential with small training batches. Each encoder stage is followed by a FiLM transformation in which scale and shift parameters are predicted from $\mathbf{p}$ by two small multilayer perceptrons. This allows the representation to be conditioned on rheological properties, enabling the network to adapt its features to flow parameters.  

A residual block at the coarsest scale forms the bottleneck before upsampling. The decoder mirrors the encoder and uses transposed convolutions ($2{\times}2$, stride 2) for upsampling. At each scale, encoder features are gated by an attention mechanism, which computes a spatial mask $\psi\in[0,1]$ from the encoder and decoder activations and modulates the skip connection as $\psi \odot \mathbf{x}_{\mathrm{enc}}$. Attention gates help suppress irrelevant features and emphasize flow pathways, improving skip fusion in complex topographies. Gated skip features and upsampled decoder activations are concatenated and passed through a residual block.  

The final decoder features feed into two output heads. A $1{\times}1$ convolution produces runout segmentation logits, which are converted to probabilities by a sigmoid during inference. To enforce physically plausible deposits, the predicted mask is thresholded at 0.5 and concatenated with decoder features before predicting thickness. A second $1{\times}1$ convolution followed by ReLU then produces the deposit thickness $\hat{h}(x)\geq 0$, ensuring that deposits remain spatially constrained within the predicted runout zone.  

The base channel width is $C_0{=}32$, which doubles at each encoder stage ($[C_0,\,2C_0,\,4C_0,\,8C_0]$) with a $16C_0$ bottleneck, mirrored in the decoder. All convolutions use $3{\times}3$ kernels unless otherwise noted, and in-place ReLU activations are applied throughout.  
 
\subsection{Training and evaluation}

The dataset was partitioned into training (80\%), validation (10\%), and test (10\%) subsets with no overlap in DEM tiles across splits. To improve invariance to map orientation and reduce overfitting, on-the-fly augmentation applied random flips and rotations to both inputs and targets.   We trained the U\mbox{-}Net with FiLM conditioning and attention gates described above, configured with eight input channels, three conditioning parameters, and a base channel width of 32. Optimization used Adam with a learning rate of $10^{-3}$ and weight decay of $10^{-5}$. A ReduceLROnPlateau scheduler (factor 0.5, patience 5 epochs) adapted the learning rate according to validation loss.  

The loss function jointly optimized segmentation and thickness prediction. Segmentation was supervised with a composite binary cross-entropy and Dice loss. Thickness was penalized with mean-squared error within the true mask and an additional out-of-mask penalty weighted by $\beta=0.1$, encouraging zero deposition outside the footprint. The total objective combined the mask and thickness terms with a relative weight of $\alpha=1.0$, coupling the tasks while enforcing physical plausibility.  

Models were trained for up to 500 epochs with early stopping after 10 epochs of non-decreasing loss. Group Normalization and dropout (0.2) within residual blocks provided additional regularization. The best-performing model, determined by minimum validation loss, was checkpointed for testing.  

At inference, the segmentation head produced logits, with probabilities obtained via a sigmoid and thresholded at 0.5 to yield binary masks. The thickness head applied a ReLU to ensure non-negativity. Model performance was evaluated on the held-out test set. For the footprint we report precision, recall, F1 score, and intersection-over-union (IoU). For thickness we report MSE and RMSE over inundated cells, together with an out-of-mask RMSE to quantify spillover.

\subsection{Computational setup}

Numerical simulations were executed as array jobs on the University of Cambridge CSD3 CPU cluster (SLURM scheduler), requiring approximately \(800\) core-hours in total and \(\sim\!30\)~hours of wall-clock time.  

Model training was conducted on a virtual machine equipped with four NVIDIA~A100 GPUs.  
For inference and benchmarking, we used a local desktop workstation equipped with an Intel~Xeon~W2\textendash2423 CPU (12~cores, 3.0~GHz), 64~GB~RAM, an NVIDIA~RTX~A4000 GPU (16~GB~VRAM), and a Samsung~PM9A1~NVMe~SSD.  
All inference times reported in the Results section were measured on this desktop configuration.

\backmatter





\bmhead{Acknowledgements}

We thank Alessandro Mondini for helpful discussions related to the model outputs.

\section*{Declarations}


\noindent
The authors declare no competing interests.\\
Data and model weights will be made available after publication.\\
Codes will be made available after publication. \\

\noindent

\end{document}